# Optimizing Ingredient Substitution Using Large Language Models to Enhance Phytochemical Content in Recipes

Luís Rita [1], Josh Southern [2], Ivan Laponogov [1], Kyle Higgins [1,3] and Kirill Veselkov [1,4]*

[1] Division of Cancer, Department of Surgery and Cancer, Faculty of Medicine, Imperial College London, London, UK

[2] Department of Computing, Faculty of Engineering, Imperial College London, London, UK

[3] Department of Neurobiology, Boston Children's Hospital and Harvard Medical School, Boston, MA, USA

[4] Department of Environmental Health Sciences, Yale University, New Haven, CT, USA

* Correspondence: kirill.veselkov04@imperial.ac.uk

**Abstract:** In the emerging field of computational gastronomy, aligning culinary practices with scientifically supported nutritional goals is increasingly important. This study explores how large language models (LLMs) can be applied to optimize ingredient substitutions in recipes, specifically to enhance the phytochemical content of meals. Phytochemicals are bioactive compounds found in plants, which, based on preclinical studies, may offer potential health benefits. We fine-tuned models, including OpenAI's GPT-3.5, DaVinci, and Meta's TinyLlama, using an ingredient substitution dataset. These models were used to predict substitutions that enhance phytochemical content and create a corresponding enriched recipe dataset. Our approach improved Hit@1 accuracy on ingredient substitution tasks, from the baseline 34.53 ± 0.10% to 38.03 ± 0.28% on the original GISMo dataset, and from 40.24 ± 0.36% to 54.46 ± 0.29% on a refined version of the same dataset. These substitutions led to the creation of 1,951 phytochemically enriched ingredient pairings and 1,639 unique recipes. While this approach demonstrates potential in optimizing ingredient substitutions, caution must be taken when drawing conclusions about health benefits, as the claims are based on preclinical evidence. Future work should include clinical validation and broader datasets to further evaluate the nutritional impact of these substitutions. This research represents a step forward in using AI to promote healthier eating practices, providing potential pathways for integrating computational methods with nutritional science.

## 1. Introduction

In recent years, computational gastronomy emerged as an interdisciplinary field, combining computational techniques with culinary science to enhance food preparation and consumption. One of the primary goals of this field is to develop methods for ingredient substitution, aimed at improving nutritional content, preserving flavor integrity, and aligning meals with specific dietary needs. A key focus has been the integration of phytochemically enriched ingredients into diets, which has shown *in silico* potential to target biological networks of chronic diseases like cancer [1], Alzheimer's disease (AD) [2], and COVID-19 [3].

Phytochemicals, bioactive compounds found in plants, have gathered significant attention due to their antioxidant, anti-inflammatory, and anti-carcinogenic properties. Preclinical studies suggest that these compounds may play a role in disease prevention and treatment. For instance, brassinolide, a phytochemical present in tea, has shown potential to inhibit tumor growth and induce apoptosis in cancer cells [4]. In the context of AD, quercetin, found in extra virgin olive oil, has been linked to improved brain health by exhibiting antioxidant and anti-inflammatory effects [5]. Moreover, genistein, a phytochemical in blackcurrant, has been investigated for its immune-supporting properties, including its potential to modulate inflammation and interfere with viral replication, making it relevant in the study of COVID-19 [6].

Initial attempts at ingredient substitution used statistical methods, such as Term Frequency–Inverse Document Frequency, to identify potential substitutes based on the occurrence patterns within large recipe datasets. These methods analysed the co-occurrences of ingredients across a large corpus of recipes to identify statistically significant substitutes [7–9]. Later, co-occurrence-based methods further refined this approach by constructing ingredient networks to map their relationships across recipes, suggesting substitutes based on their mutual presence in culinary contexts [10–14]. The introduction of language model-based methods marked a significant evolution, utilising natural language processing techniques such as word2vec [15], BERT [16], and R-BERT [17] to capture semantic relationships between ingredients [15]. This approach proved effective in improving ingredient substitution tasks through learned embeddings [18], although language models require substantial computational resources and may not always capture the full culinary context.

More recently, graph neural networks (GNNs) have been utilised to combine the relational information encoded in ingredient graphs with the specific context of given recipes, leading to a deeper understanding of ingredient interactions [18]. Large-scale graphs, such as FlavorGraph, have been introduced to explore ingredient substitutions and food pairings, demonstrating the potential of GNN-based methods in culinary applications [19]. However, success in this area relies heavily on the quality and curation of underlying graph data. Building on this approach, GISMo was introduced, a GNN-based model that incorporates both recipe-specific contexts and ingredient relationships from FlavorGraph. By constructing a benchmark dataset, Recipe1MSubs, which includes ingredient substitution pairs extracted from user comments, GISMo significantly outperforms previous methods in ranking plausible ingredient substitutions. With a performance improvement of, at least, 14% in the top substitute ingredient prediction (Hit@1) over existing models [20].

The latest stage in this evolving field is represented by LLMs, which promise to overcome the limitations of previous approaches by leveraging their capacity for understanding and generating human-like text [21,22]. The introduction of LLMs, such as GPT-3 developed by OpenAI [21], presents an approach to address the limitations of previous methods for ingredient substitution. Furthermore, while language model-based methods and GNNs represent significant advancements, they still face challenges in capturing the full culinary context and ensuring gastronomically sensible substitutions [20]. LLMs, trained on extensive and diverse culinary datasets, can potentially offer more contextually aware and accurate ingredient substitutions by leveraging their understanding of both the syntax and semantics of culinary texts [23]. This capacity for high-level language comprehension and manipulation allows for considering factors such as ingredient compatibility. Importantly, LLMs can be fine-tuned for specific tasks such as ingredient substitution [22].

Recognizing the limitations of statistical, co-occurrence, language and GNN-based methods, our research proposes a unique approach by leveraging the capabilities of LLMs for ingredient substitution. LLMs, such as OpenAI's GPT-3.5 [22], DaVinci [21], and Meta's TinyLlama [24], have demonstrated state-of-the-art performance across a range of natural language processing tasks, from text generation to semantic understanding [16,25]. By fine-tuning these models on a dataset of recipes and ingredient substitutes, we aim to develop an algorithm that not only understands the interplay of flavors and nutritional aspects in cooking but also tailors suggestions to the preferences and requirements of each user. In this paper, we benchmark our ingredient substitution algorithm against the current state of the art GISMo, to demonstrate its superiority in generating contextually appropriate ingredient substitutions [20]. We used the Hit@1 accuracy metric to benchmark our models' performance against state-of-the-art methods. After identifying phytochemically enriched substitutes, we generated a new set of recipes aimed at targeting biological networks associated with cancer, Alzheimer's disease (AD), and COVID-19 (Figure 1).

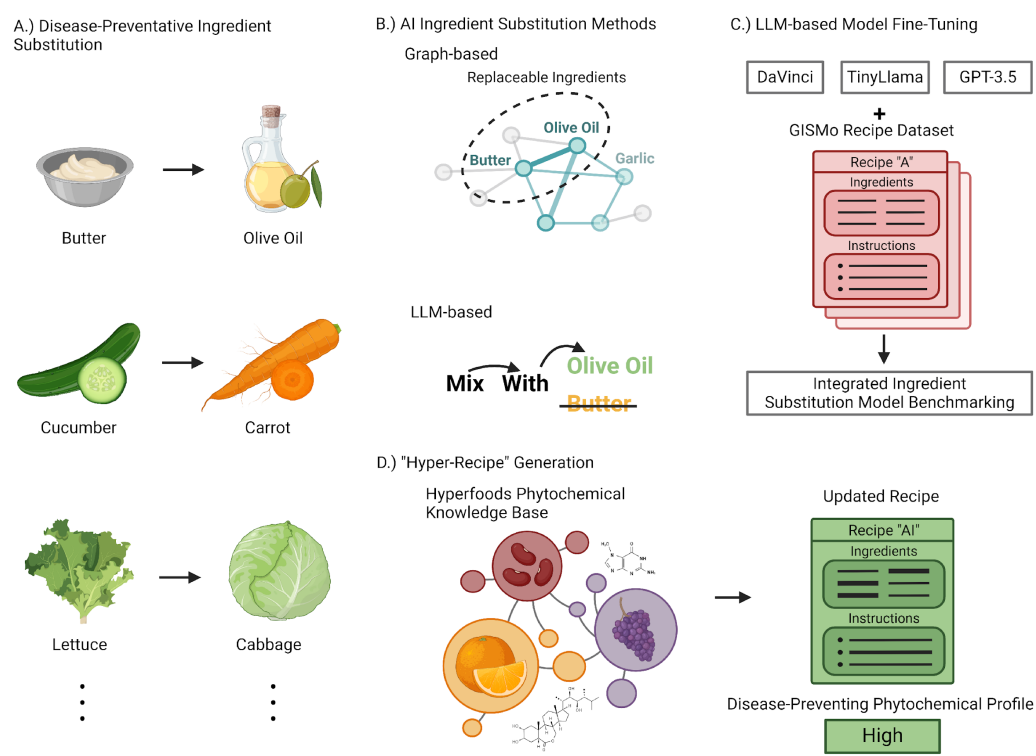

**Figure 1.** Ingredient Substitution Methods. **A)** Disease-Preventative Ingredient Substitution: it illustrates the process of substituting ingredients to enhance the phytochemical profile of recipes focused on disease-specific prevention. Examples include substituting butter with olive oil, cucumber with carrot, and lettuce with cabbage. **B)** Ingredient Substitution Methods: it compares graph-based approaches, which rely on ingredient co-occurrence and relational data, with LLM-based approaches, which utilise advanced language models for more context-sensitive substitutions. **C)** LLM-Based Model Fine-Tuning: it details the fine-tuning of LLMs such as DaVinci, GPT-3.5 and TinyLlama for the ingredient substitution task using the Recipe1MSubs dataset. It includes the benchmarking process against the GISMo model to evaluate performance improvements. **D)** Phytochemically Enriched Recipe Generation: it details the generation of recipes with enhanced phytochemical profiles tailoring diseases biological networks with the best performer model.

## 2. Materials and Methods

*2.1. Recipe and Ingredient Substitution Datasets*

Our research started with the study of Recipe1MSubs dataset, provided by Meta, containing 70,520 pairs of ingredient substitutes with the respective recipes [20], which is a subset from the Recipe1M dataset [26]. Recipe1MSubs dataset contains 49,044 designated for training, 10,729 for validation, and 10,747 for testing. Each recipe within this dataset is organized in a structured format, beginning with the recipe title, followed by the list of ingredients, with associated quantities, and, finally, the cooking instructions. The original GISMo model was trained in this dataset using a methodology focused on ingredient context and co-occurrence as the benchmark of our study.

*2.2. GISMo Benchmark*

We re-run the machine learning architecture and parameters as previously described in the reference study [20] to serve as a baseline for comparison with new models developed using LLMs. This benchmark setup includes a learning rate of 5e-05, weight decay of 0.0001, 300 hidden units per layer, an embedding dimension of 300, and a dropout rate of 0.25, across two layers, and involves 400 training rounds with regular negative sampling and random embedding initialization. Unlike data augmentation, ingredient sets are included to enhance the model's context sensitivity without altering the original dataset's composition. Average pooling is utilised for contextual embedding. By replicating this validated model configuration, we establish a standard against which the performance of newer, potentially more sophisticated LLM-based models can be evaluated, ensuring that any improvements in ingredient substitution accuracy are attributable to the capabilities of LLMs. To augment the capabilities of the GISMo model, we introduced several enhancements intended to improve the accuracy of ingredient substitution predictions, namely by considering each ingredients' food categories.

*2.3. Incorporation in GISMo of Food Category Feature*

Using GPT-4-0613, the latest of OpenAI's language models, we categorize ingredients into predefined culinary groups. This process involved a Python script utilising the *pandas* library for dataset manipulation and the *openai* library for API interactions. A function, *categorize_ingredient*, was used to query GPT-4-0613 with each ingredient, requesting its classification into one of 23 categories ranging from common food groups like *Fruits* and *Vegetables* to more specialized ones such as *Confectioneries* and *Aquatic foods* (Appendix A). By setting the *temperature* parameter to 0, the script prioritises reproducibility to ensure consistency in GPT-4-0613's responses. This approach processed a CSV file of ingredients, appending a *category* column with the GPT-4-0613-determined categories to the dataset. The augmented dataset, saved as a new CSV file, now serves as a tool for ingredient substitution models, enabling more contextually relevant substitutions.

*2.4. Dataset Filtration Based on Substitution Validity*

To enhance the ingredient substitution model's accuracy, we used GPT-3.5-Turbo with an asynchronous Python script for evaluating the validity of proposed ingredient substitutions. This process involved sending detailed prompts to GPT-3.5-Turbo, asking if one ingredient could feasibly substitute another within a specific recipe, and classifying responses into *Correct*, *Potential*, or *Incorrect* to determine their suitability. By processing substitutions in multiple batches using the *aiohttp* library for asynchronous HTTP requests, we efficiently assessed the 70,520 substitutions, thereby accelerating the evaluation process. Substitutions categorized as *Correct* were considered suitable and retained, *Potential* indicated possible suitability requiring further consideration, and *Incorrect* were considered inappropriate, leading to their removal from the dataset. The final results, saved into a JSON file, formed a filtered dataset for retraining the model, ensuring it is based on accurate substitution data. Key settings included a prediction temperature of 0.5, a limit of 10 output tokens, and five runs to evaluate prediction stability. Plus, a batch size of 100 substitutions with respective recipes to avoid reaching the maximum number of requests per second.

*2.5. Fine-tuning Language Models for Substitution Predictions*

We used GPT-3.5-Turbo-1106, DaVinci-002, and TinyLlama-1.1B to predict viable ingredient substitutions, fine-tuning each with consistent specifications to ensure comparability. Key settings included a prediction temperature of 0.5, a limit of 10 output tokens, and five runs to evaluate prediction stability, all conducted over a single epoch.

The fine-tuning process for TinyLlama-1.1B models in our experimental configuration, we refined our model's fine-tuning process with selected hyperparameters encapsulated within the *TrainingArguments* setup. This configuration specified an output directory, a per-device train batch size of 8 (due to memory constraints) and applied gradient accumulation over 4 steps to efficiently balance between computational demand and memory constraints. The model optimisation was conducted using *paged_adamw_32bit* with a learning rate set at 5e-4, and a cosine learning rate scheduler was employed for optimal learning rate adjustments throughout the training phase. A save strategy based on epochs was utilised, coupled with logging and evaluation intervals set at 25 and 50 steps respectively, aligning with an evaluation strategy that triggers at specified steps to closely monitor the model's performance. The training was streamlined to complete within 1 epoch to ensure quick adaptation while preventing overfitting, without setting a maximum step limit and avoiding mixed precision training to maintain computational accuracy. The *SFTTrainer* was used in the training process, directly interfacing with the training and validation datasets, and was configured with *peft_config* for tailored pre-fine-tuning adjustments. Text preprocessing was managed using a specified *dataset_text_field* and *tokenizer*, with packing disabled and a maximum sequence length of 512 to standardize input data handling. This approach aimed at enhancing the model's learning efficiency, prioritizing a balance between computational resource optimisation and achieving high-quality model training.

Building upon the filtration methodology outlined above, we randomly chose one of the filtered datasets and fine-tuned four final models considering only the *Correct* substitutions to further refine the accuracy of predictions. TinyLlama-1.1B, DaVinci-002 and GPT-3.5-Turbo-1106 and GISMo models were fine-tuned incorporating these high-quality substitutions.

Training samples were provided in prompt completion format for DaVinci-002 and TinyLlama-1.1B, and chat completion format for GPT-3.5-Turbo-1106. Number of epochs, training steps and batch sizes chosen are in Appendix B.

## 2.6. Evaluation of Ingredient Substitution Accuracy

To validate the accuracy of ingredient substitution predictions generated by LLMs, we developed an algorithm to standardize and process ingredient names before comparing them to a ground truth dataset derived from the Recipe1M dataset. We begin by extracting predictions from the model output, where each line contains an original ingredient, its corresponding ground truth substitute, and the predicted substitution. To ensure consistency across ingredient names, several preprocessing steps are applied, including converting all text to lowercase, removing numeric values, and applying predefined rules to replace or eliminate special characters. This normalization was intended to maintain uniformity in ingredient representation. After preprocessing, a clustering mechanism is used to group similar ingredients, accounting for variations in lexical forms such as singular and plural versions or different types of the same ingredient (e.g., basmati rice and long grain rice). Each ingredient is assigned a unique cluster identifier to ensure that similar ingredients are treated as equivalent during comparison.

Once the LLM predicted ingredient names are uniformized and categorized, the core of the evaluation involves comparing the predicted substitutions against the ground truth using the Hit@1 metric. This metric assesses the model's precision by determining whether the first predicted substitution matches the ground truth or falls within the same ingredient cluster. For example, if the ground truth substitution is *barley* and the model predicts *basmati rice*, both ingredients would be considered correct if they belong to the same grain cluster. Hit@1 focuses on measuring the accuracy of the model's top recommendation, as this is the most critical in real-world applications where users often act on the first suggestion. By prioritizing precision in the initial substitution, Hit@1 provides a measure of the LLM's ability to generate viable and contextually appropriate ingredient substitutions.

## 2.7. Phytochemically Enriched Recipe Generation

Finally, we integrated phytochemically enriched ingredients based on their ability to target molecular networks responsible for disease development in cancer [1], AD [2], and COVID-19 [3]. By applying the best-performing model from our comparative analysis, we substituted all ingredients across our dataset with alternatives that elevate the content of targeted phytochemicals. Recipes were then evaluated and ranked based on their cumulative phytochemical profile. Only salads were considered given the lower number of cooking processes involved in their preparation and, consequently, higher chances of phytochemical preservation [27] (Figure 2).

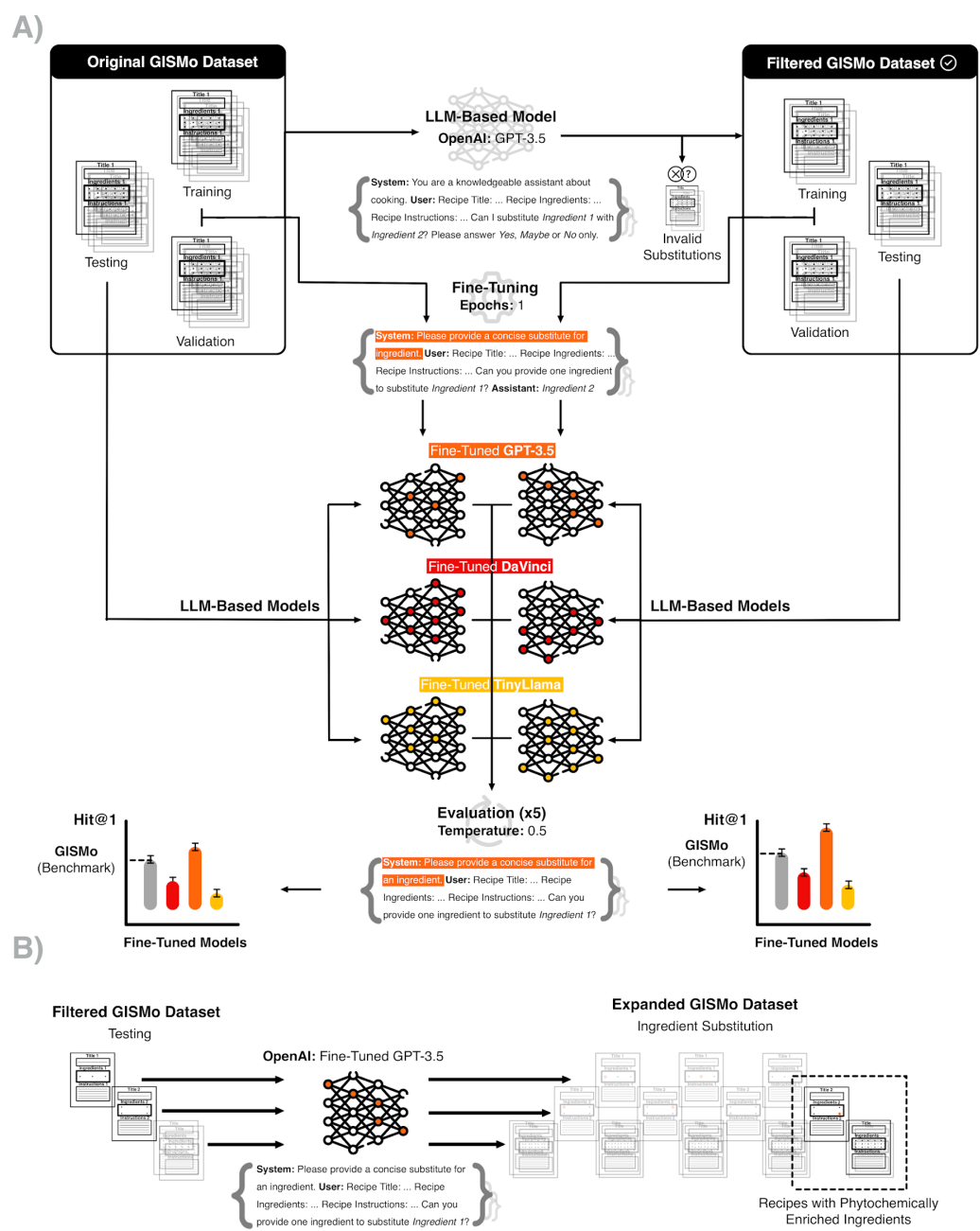

**Figure 2.** This figure is divided into two parts, A and B, detailing the processes involved in our study. **A)** Fine-tuning process of three LLMs—OpenAI's GPT-3.5, DaVinci, and Meta's TinyLlama—benchmarked against the state-of-the-art GISMo ingredient substitution model. The process involved prompt engineering to optimize model responses for providing single ingredient substitutions. It includes visual representations of workflow diagrams showcasing the application of prompts and evaluation of model responses. The evaluation of fine-tuned models is depicted, highlighting the significant improvement in Hit@1 accuracy compared to the GISMo benchmark. **B)** Application of the best-performing LLM to generate new recipes. It illustrates the creation of new culinary recipes with the highest number of phytochemically enriched ingredients. The process from model selection, recipe generation, to the final selection based on phytochemical content is demonstrated.

## 3. Results

The original Recipe1MSubs dataset contains 70,520 ingredient substitutions, with 49,044 designated for training, 10,729 for validation, and 10,747 for testing. Filtering this dataset through five separate runs with GPT-3.5-Turbo-1106 resulted in five filtered sets, with an average of 31,819 ± 67 training samples, 7,094 ± 25 validation samples, and 7,085 ± 21 testing samples, totaling 45,998 ± 85 (Appendix C). From these, a randomly selected dataset with 44,615 ingredient substitutions, divided into 31,063 for training, 6,831 for validation, and 6,721 for testing, was used to rerun GISMo and fine-tune LLMs as detailed below.

Running the GISMo model five times with both the original and filtered datasets yielded Hit@1 accuracies of 34.53 ± 0.10% and 40.24 ± 0.36%, respectively. Integrating food categories (Appendix A) as additional node features into our computational model with the original dataset resulted in a Hit@1 value of 34.62% (Appendix D).

Subsequent experiments with fine-tuned LLMs on the original GISMo ingredient substitution dataset yielded Hit@1 values of 20.09 ± 0.31% for DaVinci-002, 20.93 ± 0.29% for TinyLlama-1.1B, and 38.03 ± 0.28% for GPT-3.5-Turbo-1106. Using the filtered dataset in conjunction with these fine-tuned models resulted in improved Hit@1 values of 29.43 ± 0.30%, 34.53 ± 0.32%, and 54.46 ± 0.29%, respectively (Appendix D).

This process resulted in the generation of 1,951 phytochemically enriched ingredient substitutions, leading to 1,639 unique recipes, each featuring at least one phytochemically rich ingredient. As an example, the *Watercress Salad* was enriched with phytochemicals predicted *in silico* to have potential relevance for COVID-19, the *Kale and Quinoa Salad* was enhanced with compounds associated *in silico* with AD and COVID-19, and the *Thai-Style Beef Salad* was optimized with ingredients *in silico* predicted to target cancer, COVID-19, and AD (see Appendix E for details).

**Table 1** Performance of Fine-Tuned Models. Performance evaluation of fine-tuned models DaVinci-002, TinyLlama-1.1B, GPT-3.5-Turbo-1106, and GISMo, tested on both the original and filtered Recipe1MSubs datasets. The displayed Hit@1 accuracies are presented as mean ± standard deviation. The models are ordered in the table from best to worst performance based on their Hit@1 accuracy.

| Recipe1MSubs Dataset | Fine Tuned Model | Hit@1 (%) |
| --- | --- | --- |
| Filtered | **GPT-3.5-Turbo-1106** | **54.46 ± 0.29** |
| | GISMo | 40.24 ± 0.36 |
| | TinyLlama-1.1B | 34.53 ± 0.32 |
| | DaVinci-002 | 29.43 ± 0.30 |
| Unfiltered | **GPT-3.5-Turbo-1106** | **38.03 ± 0.28** |
| | GISMo | 34.55 ± 0.11 |
| | TinyLlama-1.1B | 20.93 ± 0.29 |
| | DaVinci-002 | 20.09 ± 0.31 |

## 4. Discussion

*4.1. Incorporation in GISMo of Food Category Feature*

An initial strategy we explored was the enhancement of the GISMo model through the incorporation of an additional node feature—food categories for each ingredient, classified into one of the 23 categories utilised in FooDB, based on classifications retrieved via the GPT-4. However, contrary to our expectations, this modification did not yield any improvements in the model's performance. This outcome may be attributed to several factors. Firstly, including this additional categorical information might have led to overfitting the model to the training data, compromising its ability to generalize effectively to unseen data in the test set. Additionally, another potential reason could be that part of the value of ingredient categorization might have been indirectly achieved by the model's consideration of ingredient co-occurrence in recipes alongside the presence of flavor molecules. These inherent features within the training data might already provide a basis for the model to make substitution predictions without the need for explicit categorical labels.

*4.2. Dataset Filtration Based on Substitution Validity*

With the goal of optimizing ingredient substitution, our study introduced an improvement by integrating the capabilities of GPT-3.5 with the GISMo model. While GISMo independently showcased a threefold enhancement in performance compared to prior methods [28], our approach to refine the GISMo model through the preliminary filtration of the original dataset via GPT-3.5's API further increased this improvement. This filtration process involved the exclusion of *Potential* and *Incorrect* substitutions from the dataset, thereby ensuring a higher quality of data for model training and application.

The filtration step encompassed five different datasets, and although one was randomly selected to rerun GISMo, the improved results are generalizable across all due to their almost perfect ingredient substitute similarity across training, validation, and testing datasets. To demonstrate the consistency of our filtration process, here are examples of substitutions that were consistently classified across the five runs: **A)** *Correct Substitutions*: orange juice to pineapple juice, carrot to red pepper, black bean to chickpea, basil to dried oregano, onion to shallot. **B)** *Potential Substitutions*: lemon to orange, apple to peach, apple to apricot, water to wine, blueberry to strawberry. **C)** *Incorrect Substitutions*: seedless watermelon to lime, fresh cilantro to ground coriander, horseradish to honey, carrot to seasoning salt, clove to garlic.

*4.3. LLM Fine-Tuning for Ingredient Substitution*

Using Recipe1MSubs dataset, our experiments explored the benefits of fine-tuning DaVinci, TinyLlama and GPT-3.5. The first two models did not demonstrate any performance enhancements over the initial method. In contrast, the fine-tuned model leveraging the GPT-3.5 showed a 4% improvement in performance over the GISMo model. Building upon this Recipe1MSubs filtered dataset, we ventured to fine-tune the same three models. Again, the GPT-3.5 model was the only one showing an increase in performance (20%) when compared with current state of the art.

The findings of this study underscore the importance of data quality and model compatibility in the development of ingredient substitution algorithms. The superior performance achieved through the combination of GPT-3.5's advanced language processing capabilities and the GISMo model's framework highlights the potential of leveraging state-of-the-art AI technologies to refine and enhance existing computational models.

*4.4. Phytochemically Enriched Recipe Generation*

We specifically selected examples of recipes with ingredients phytochemically enriched targeting COVID-19; COVID-19 and AD; and COVID-19, AD and cancer molecular networks. In the previous order, those are *Watercress Salad*, *Kale and Quinoa Salad* and *Thai-Style Beef Salad* (Appendix E). We exclusively considered salads in this analysis due to their minimal food processing steps. This choice was made because fewer processing steps generally help preserve the phytochemicals with the health benefits discussed. Salads undergo minimal thermal processing, which helps maintain the integrity of essential nutrients and active compounds compared to more extensively cooked dishes [27].

*4.6. Limitations*

One inherent limitation is the diversity of the training datasets used to fine-tune the LLMs. Although these datasets are extensive, they may not fully capture the vast diversity of global cuisines and dietary preferences, potentially impacting the model's ability to generalize across different culinary traditions and suggest culturally and regionally appropriate substitutions. Additionally, the methodology primarily focuses on textual data, which might not capture the full spectrum of culinary contexts, including taste profiles, textures, and the interplay of flavors. LLMs, while proficient in parsing and generating text, have limited capacity to understand and replicate the sensory experiences of cooking and eating.

Additionally, the fine-tuning process, especially when using a limited set of high-quality substitutions, poses a risk of overfitting, where models may become overly specialized to the training data and less capable of generalizing to unseen recipes or ingredients.

Furthermore, the reliance on the Hit@1 metric, while providing a clear measure of the model's ability to suggest the correct first substitution, does not capture the overall utility and flexibility of the model in providing a range of suitable alternatives.

Finally, the computational resources required for fine-tuning and deploying LLMs may also limit the accessibility of these advanced tools to researchers and practitioners with limited resources.

## 5. Conclusions

By integrating state-of-the-art LLMs like OpenAI's GPT-3.5 and DaVinci, and Meta's TinyLlama, we have developed a method that enhances the accuracy and relevance of ingredient substitutions, producing recipes with phytochemically enriched ingredients. The fine-tuning of these models on Recipe1MSubs dataset allowed us to exceed the capabilities of traditional statistical and network-based methods, achieving a significant improvement in the Hit@1 metric in the original unfiltered and filtered datasets using fine-tuned GPT-3.5. Furthermore, by incorporating specific *in silico* phytochemical data, our approach led to the generation of phytochemically enriched ingredient substitutions and recipes targeting at least one of the following disease networks - COVID-19, AD and cancer. As we continue to refine the models and expand our datasets, we anticipate that incorporating domain-specific knowledge, such as clinical and biochemical data, will be crucial for enhancing the accuracy and relevance of ingredient substitutions. Future research should focus on rigorous validation of these substitutions through clinical trials and controlled dietary studies to assess their efficacy in improving health outcomes. These developments hold promise for revolutionizing personalized nutrition and optimizing dietary practices in a scientifically robust and clinically validated manner.

**Author Contributions:** Conceptualization, K.V. and I.L.; methodology, K.V., I.L., L.R. and J.S.; investigation, L.R. and J.S.; data curation, L.R.; writing—original draft preparation, L.R., J.S. and K.H.; writing—review and editing L.R., J.S. and K.H.; visualization, L.R. and K.H.; supervision, K.V. and I.L.; project administration, K.V. and I.L.; funding acquisition, K.V. All authors have read and agreed to the published version of the manuscript.

**Funding:** This research was funded by Fundação para a Ciência e a Tecnologia, grant number 2021.05460.BD; ERC Consolidator, 724228 (LEMAN); ERC Proof of Concept, 899932 (Hyperfoods); UK Research and Innovation, 10058099; European Union, 101095359; and Vodafone Foundation, CORONA-AI/DRUGS DreamLab. We also acknowledge the large citizen science community that made the discovery of phytochemicals possible through the use of the DreamLab app.

**Data Availability Statement:** Fine-tuned models GISMo and TinyLlama, along with the scripts used for filtering the Recipe1MSubs dataset and fine-tuning the GISMo, DaVinci, GPT-3.5, and TinyLlama models, are hosted on the Bitbucket repository. Additionally, data on the categories each ingredient from Recipe1M belongs to, as well as the script used to retrieve these categories, are also made available. The repository can be accessed at the following URL: https://bitbucket.org/iAnalytica/recipe-generator. However, due to the use of proprietary models by OpenAI and limitations set by third-party data sources, some restrictions apply to the direct sharing of model weights and specific algorithmic implementations.

**Appendix A**

All ingredients in the Recipe1MSubs dataset were classified into one of the following 23 food categories as identified in the FooDB database [29]. The categories are:

1. Herbs and Spices
2. Fats and Oils
3. Unclassified
4. Baby Foods
5. Snack Foods
6. Dishes
7. Baking Goods
8. Confectioneries
9. Eggs
10. Milk and Milk Products
11. Animal Foods
12. Aquatic Foods
13. Beverages
14. Cocoa and Cocoa Products
15. Soy
16. Coffee and Coffee Products
17. Gourds
18. Teas
19. Pulses
20. Cereals and Cereal Products
21. Nuts
22. Fruits
23. Vegetables

## Appendix B

Training configurations for DaVinci-002, GPT-3.5-Turbo-1106, and TinyLlama-1.1B, as well as their filtered dataset variants, are detailed below. This table includes the number of epochs, training steps, and batch sizes used for each model. Models highlighted with an (*) had their training parameters manually optimized to enhance performance.

| Model | Epochs | Steps | Batch Size |
|---|---|---|---|
| DaVinci-002 | 1 | 1533 | 32 |
| DaVinci-002 (Filtered) | 1 | 1554 | 20 |
| GPT-3.5-Turbo-1106 | 1 | 1533 | 32 |
| GPT-3.5-Turbo-1106 (Filtered) | 1 | 1554 | 20 |
| TinyLlama-1.1B-1.1B* | 1 | 1532 | 8 |
| TinyLlama-1.1B-1.1B (filtered)* | 1 | 970 | 8 |

## Appendix C

Performance evaluation of the GISMo model, after being trained, validated, and tested on five datasets generated by filtering the original dataset through the GPT-3.5-Turbo-1106. The table below details the Hit@1 accuracy metric and the number of recipes used in each phase—training, validation, and testing—across all runs.

| Filtering Run | Hit@1 (%) | Training | Validation | Testing | Total |
|---|---|---|---|---|---|
| Run 1 | 40.28 | 31,733 | 7,082 | 7,080 | 45,895 |
| Run 2 | 40.82 | 31,795 | 7,097 | 7,056 | 45,948 |
| Run 3 | 40.21 | 31,797 | 7,083 | 7,096 | 45,976 |
| Run 4 | 39.98 | 31,908 | 7,073 | 7,113 | 46,094 |
| Run 5 | 39.90 | 31,860 | 7,136 | 7,081 | 46,077 |
| Final | 40.24 ± 0.36 | 31,819 ± 67 | 7,094 ± 25 | 7,085 ± 21 | 45,998 ± 85 |

## Appendix D

Comparative analysis of three models—DaVinci-002, GPT-3.5-Turbo-1106, and TinyLlama-1.1B—evaluating their performance on the Recipe1MSubs dataset using the Hit@1 accuracy metric across multiple runs. The analysis includes descriptions of the datasets used.

Dataset Descriptions:

- Original Dataset: 70,520 ingredient substitutions, with 49,044 for training, 10,729 for validation, and 10,747 for testing.

- Filtered Dataset: 44,615 ingredient substitutions, with 31,063 for training, 6,831 for validation, and 6,721 for testing.

The table below shows the Hit@1 accuracy for each model across five runs, along with the final average and standard deviation. In bold the best performer fine-tuned models with the original and filtered datasets. The models are ordered in the table from best to worst performance based on their Hit@1 accuracy.

| Recipe1MSubs Dataset | Fine-Tuned Model | Run 1 | Run 2 | Run 3 | Run 4 | Run 5 | Final Hit@1 (%) |
|---|---|---|---|---|---|---|---|
| Fitered | GPT-3.5-Turbo-1106 (Filtered Dataset) | 54.05 | 54.77 | 54.69 | 54.40 | 54.37 | 54.46 ± 0.29 |
| | GISMo (Filtered Dataset) | 40.28 | 40.82 | 40.21 | 39.98 | 39.90 | 40.24 ± 0.36 |
| | TinyLlama-1.1B (Filtered Dataset) | 35.07 | 34.22 | 34.46 | 34.53 | 34.37 | 34.53 ± 0.32 |
| | DaVinci-002 (Filtered Dataset) | 28.97 | 29.59 | 29.61 | 29.27 | 29.70 | 29.43 ± 0.30 |
| Unfiltered | GPT-3.5-Turbo-1106 | 38.08 | 38.25 | 37.96 | 38.28 | 37.59 | 38.03 ± 0.28 |
| | GISMo | 34.55 | 34.42 | 34.68 | 34.54 | 34.45 | 34.53 ± 0.10 |
| | TinyLlama-1.1B | 20.44 | 20.78 | 20.35 | 20.18 | 20.16 | 20.38 ± 0.25 |
| | DaVinci-002 | 20.39 | 19.73 | 19.77 | 20.29 | 20.26 | 20.09 ± 0.31 |

#### Appendix E

Three examples of recipes showcasing substitutions that account for the dish's flavor profile but also its nutritional value using phytochemically enriched ingredients:

- *Thai Style Beef Salad*: This recipe includes substitutions such as replacing mung bean sprouts with cabbage to increase the recipe's content of glucosinolates, known for their cancer-preventive properties. The use of olive oil instead of sesame oil increases the content of healthy fats and antioxidants, supporting cognitive health and cardiovascular health.

- *Super Corn Salad*: substitutions in this recipe include using olive oil instead of vegetable oil to provide healthier monounsaturated fats and antioxidants. Carrots replace pepper to increase the beta-carotene content, beneficial for immune function, and dill is used instead of tarragon, providing a different set of phytonutrients beneficial for inflammation reduction.

- *Pineapple-Cabbage Salad*: In this recipe, radishes are substituted with carrots to increase the beta-carotene content, and peas are replaced with more carrots to further enhance the dish's vitamin A content, which is crucial for immune system function and vision.

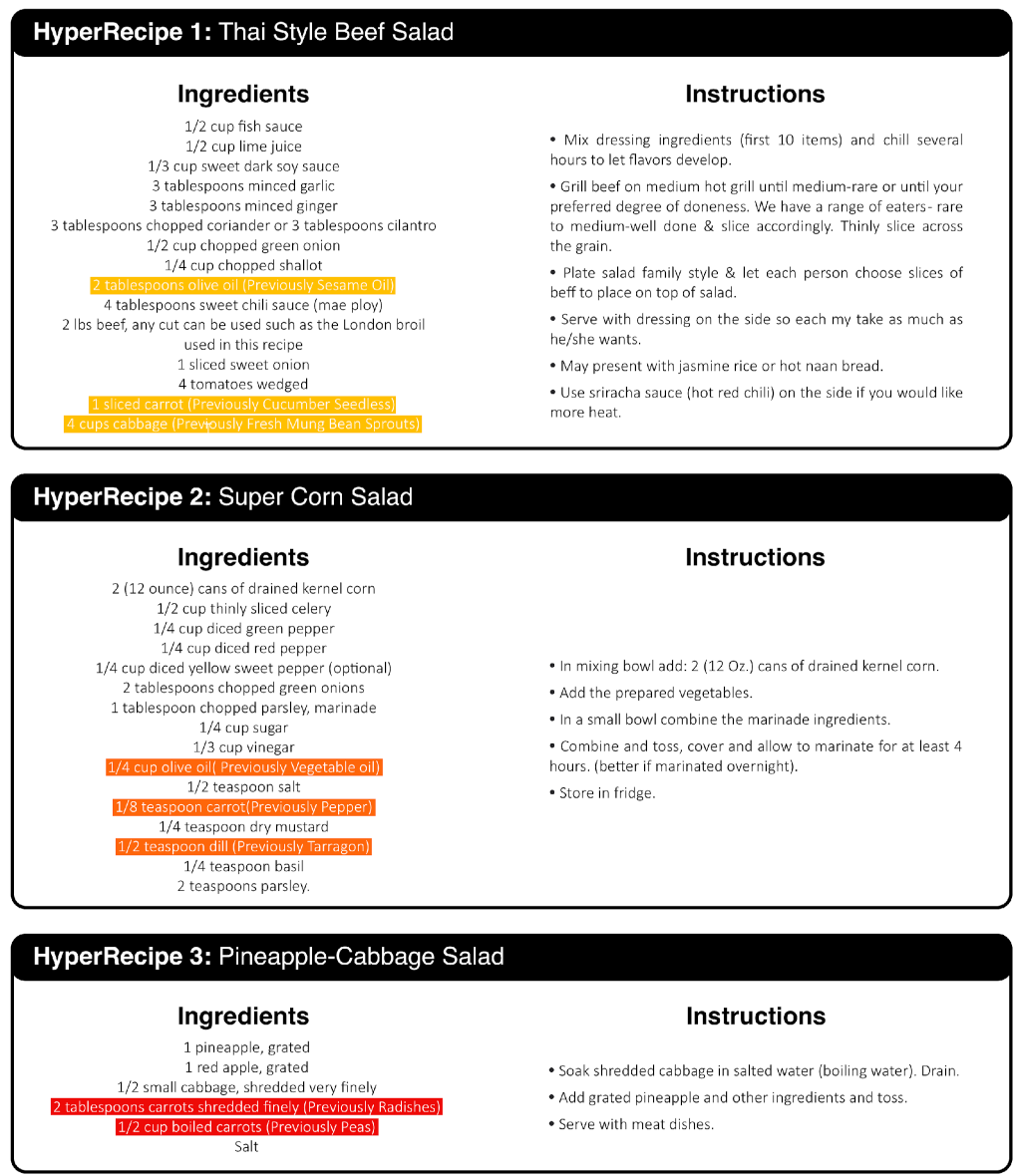

Figure 3. Phytochemically Enriched Recipes. Three examples showing the old and substituted ingredients considering the whole context of each recipe.